\documentclass{article} 
\usepackage{nips11submit_e,times}
\usepackage{amsmath,amsfonts,amssymb}

\newtheorem{theorem}{Theorem}

\newcommand{\beq}{\begin{equation}}
\newcommand{\eeq}{\end{equation}}

\newcommand{\beqa}{\begin{eqnarray}}
\newcommand{\eeqa}{\end{eqnarray}}

\newcommand{\beqan}{\begin{eqnarray*}}
\newcommand{\eeqan}{\end{eqnarray*}}

\newcommand{\eqdef}{\stackrel{\rm def}{=}}

\newlength{\minipagewidth}
\newcommand{\bookbox}[1]{
\par\medskip\noindent
\framebox[0.83\textwidth]{
\begin{minipage}{\minipagewidth}
{#1}
\end{minipage} } \par\medskip }

\newtheorem{definition}{Definition}

\renewcommand{\epsilon}{\varepsilon}
\renewcommand{\leq}{\leqslant}
\renewcommand{\geq}{\geqslant}
\renewcommand{\hat}{\widehat}

\newcommand{\Esp}{\mathbb{E}}
\newcommand{\cA}{\mathcal{A}}
\newcommand{\cB}{\mathcal{B}}

\newcommand{\cG}{\mathcal{G}}

\newcommand{\cS}{\mathcal{S}}

\newcommand{\cX}{\mathcal{X}}
\newcommand{\cO}{\mathcal{O}}
\newcommand{\cP}{\mathcal{P}}
\newcommand{\cI}{\mathcal{I}}
\newcommand{\cR}{\mathcal{R}}

\newcommand{\cJ}{\mathcal{J}}

\newcommand{\Nat}{\mathbb{N}}
\newcommand{\Hist}{\mathcal{H}}

\newcommand{\argmax}{\mathop{\mathrm{argmax}}}

\title{Selecting the State-Representation \\
in Reinforcement Learning}

\author{
Odalric-Ambrym Maillard
\\
INRIA Lille\\ \texttt{odalricambrym.maillard@gmail.com} \\
\And
R\'emi Munos \\
INRIA Lille \\
\texttt{remi.munos@inria.fr} \\
\AND
Daniil Ryabko \\
INRIA Lille \\
\texttt{daniil@ryabko.net} 
}

\nipsfinalcopy 

\begin{document}

\maketitle

\begin{abstract}
The problem of selecting the right state-representation in a reinforcement learning problem is considered. 
Several models (functions mapping past observations to a finite set) of the observations are given, and it is known 
that for at least one of these models the resulting state dynamics are indeed Markovian. Without
knowing neither which of the models is the correct one, nor what are the probabilistic characteristics of the resulting 
MDP, it is required to obtain as much reward as the optimal policy for the correct model (or for the best of the correct models, if there are several). We propose an algorithm that achieves that, 
with a regret of order $T^{2/3}$ where $T$ is the horizon time.
\end{abstract}

\section{Introduction}
We consider the problem of selecting the right state-representation in an average-reward reinforcement learning problem. 
Each state-representation is defined by a model $\phi_j$ (to which corresponds a state space $\cS_{\phi_j}$) and we assume that the 
number $J$ of available models is finite and that (at least) one model 
is a weakly-communicating Markov decision process (MDP). 
We do not make any assumption at all about the other models. This problem is considered in the general reinforcement learning setting,
 where an agent interacts with an unknown environment in a single stream of repeated observations, actions and rewards.
There are no ``resests,'' thus all the learning has to be done online. 
Our goal is to construct an algorithm that performs almost as well as the algorithm that knows both which model is a MDP (knows the ``true'' model) and the characteristics of this MDP (the transition probabilities and rewards).

Consider some examples that help motivate the problem. The first example is high-level feature selection. Suppose 
that the space of histories is huge, such as the space of video streams or that of game plays. In addition to these data,
we also have some high-level features extracted from it, such as ``there is  a person present in the video'' or 
``the adversary (in a game) is aggressive.'' We know that most of the features are redundant, but we also know that 
some combination of some of the features describes the problem well and exhibits Markovian dynamics. Given a potentially large
number of feature combinations of this kind, we want to find a policy whose average reward is as good as that of the best policy
for the right combination of features.  Another example is  bounding the order of an MDP. 
The process is known to be $k$-order Markov, where  $k$ is unknown but un upper bound $K >>k$  is given.
The goal is to perform as well as if we knew $k$.
Yet another example is selecting the right discretization. The environment is an MDP with a continuous state space. 
We have several candidate quantizations of the state space, one of which gives an MDP. Again, we would like to find a policy that is as good 
as the optimal policy for the right discretization. This example also opens the way for extensions of the proposed approach: we would like 
to be able to treat an infinite set of possible discretization, none of which may be perfectly Markovian. The present work can be considered 
the first step in this direction.

It is important to note that we do not make any assumptions on the ``wrong'' models (those that do not have Markovian dynamics). 
Therefore, we are not able to {\em test} which model is Markovian in the classical statistical sense, since in order to do that 
we would need a viable alternative hypothesis (such as, the model is not Markov but is $K$-order Markov). In fact, the constructed
algorithm never ``knows'' which model is the right one; it is ``only'' able to get the same average level of reward as if it knew.

{\bf Previous work.}
This work builds on previous work on learning average-reward MDPs. 
Namely, we use in our algorithm as a subroutine the algorithm UCRL2 of \cite{UCRL2} 
that is designed to provide finite time bounds for undiscounted MDPs. Such a problem has been pioneered in the reinforcement learning literature by \cite{Kearns2002} and then improved in various ways by \cite{Brafman03,Strehl2006,Tewari07optimisticlinear,UCRL2,Bartlett2009};
UCRL2 achieves a regret of the order $D T^{1/2}$ in any weakly-communicating MDP with diameter $D$, with respect to the best policy for this MDP. The diameter $D$ of a MDP is defined in \cite{UCRL2} as the expected minimum time required to reach any state starting from any other state.
A related result is reported in \cite{Bartlett2009}, which improves on constants related to the characteristics of the MDP.

A similar approach has been considered in \cite{Ryabko2008}; the difference is that in that work 
the probabilistic characteristics of each model are completely known, but the models are not assumed 
to be Markovian, and belong to a countably infinite (rather than  finite) set. 

The problem we address can be also viewed as a generalization of the bandit problem (see e.g.~
\cite{ro52,LaiRo85,Auer02}): there are finitely many ``arms'', corresponding to the policies used in each model, and one of the arms is the best, in the sense that the corresponding model is the ``true'' one. In the usual bandit setting, the rewards are assumed to be i.i.d.~thus one can estimate the mean value of the arms while switching arbitrarily from one arm to the next (the quality of the estimate only depends on the number of pulls of each arm). However, in our setting, estimating the average-reward of a policy requires playing it \textit{many times consecutively}. This can be seen as a bandit problem with dependent arms, with complex costs of switching between arms.

{\bf Contribution.}
We show that despite the fact that the true Markov model of states 
 is unknown and that
nothing is assumed on the wrong representations, 
it is still possible to derive a finite-time analysis
of the regret for this problem. This is stated in Theorem~\ref{thm:main};
the bound on the regret that we obtain is of order $T^{2/3}$.

The intuition is that if the ``true'' model $\phi^*$ is known, but its probabilistic properties are not, then we still know  that there exists an optimal control policy that depends on the observed state $s_{j^*,t}$ only. 
Therefore,  the optimal rate of rewards can be obtained by a clever exploration/exploitation strategy, such as UCRL2 algorithm \cite{UCRL2}. Since we do not know in advance which model is a MDP, we need to explore them all, for a sufficiently long time 
 in order to estimate the rate of rewards that one can get using a good policy in that model.

{\bf Outline.}
In Section~\ref{sec:notations} we introduce the precise notion of model and set up the notations.
Then we present the proposed algorithm in Section~\ref{sec:main}; it uses   UCRL2 of \cite{UCRL2} as a subroutine
and selects the models $\phi$ according to a penalized empirical criterion.
 In Section~\ref{sec:disc} we discuss some directions for further development.
Finally, Section~\ref{sec:proof} is devoted to the proof of Theorem~\ref{thm:main}.

\vspace{-3mm}
\section{Notation and definitions}\label{sec:notations}

\vspace{-3mm}
We consider a space of observations $\cO$, a space of actions $\cA$, and a space of rewards $\cR$ (all assumed to be  Polish).
 Moreover, we  assume that $\cA$ is of finite cardinality $A\eqdef|\cA|$ and that $0 \in \cR \subset [0,1]$. 
The set of histories up to time $t$ for all $t \in \Nat\cup\{0\}$ will be denoted by $\Hist_{<t} \eqdef \cO\times(\cA\times\cR\times\cO)^{t-1}$, and we define the set of all possible histories by $\displaystyle{\Hist \eqdef \bigcup_{t=1}^{\infty} \Hist_{<t}}$.

\vspace{-3mm}
{\bf Environments.}
 For a Polish $\cX$, we Denote by $\cP(\cX)$ the set of probability distributions over $\cX$.
Define an environment to be a mapping from the set of histories $\Hist$ to the set of functions that map any action $a\in\cA$ to 
a probability distribution $\nu_a \in \cP(\cR\times\cO)$ over the product space of rewards and observations. 

We consider the problem of reinforcement learning when the learner interacts with 
 some \textit{unknown} environment $e^\star$. 
The interaction is sequential and goes as follows: first some $h_{<1}= \{o_0\}$ is generated according to $\iota$, 
then at time step $t>0$, the learner choses an action $a_t\in\cA$ according to the current history $h_{<t} \in\Hist_{<t}$. 
Then a couple of reward and observations $(r_t,o_t)$ is drawn according to the distribution $(e^\star(h_{<t}))_{a_t} \in \cP(\cR\times\cO)$.
Finally, $h_{<t+1}$ is defined by the concatenation of $h_{<t}$ with $(a_t,r_t,o_t)$. With these notations, at each time step $t>0$, 
$o_{t-1}$ is the last observation given to the learner before choosing an action, $a_t$ is the action output at this step, and $r_t$ is the immediate reward received after playing $a_t$.

{\bf State representation functions (models).}
Let $\cS\subset\Nat$ be some finite set; intuitively, this has to be considered as a set of states. 
A \textit{state representation} function $\phi$ is a function from the set of histories $\Hist$ to $\cS$.
For a state representation function $\phi$, we will use the notation  $\cS_{\phi}$ for its set of states, 
and   $s_{t,\phi}:=\phi(h_{<t})$.

In the sequel, when we talk about a Markov decision process, it will be assumed to be \textit{weakly communicating}, which means that for each pair of states $u_1,u_2$ there exists $k\in\Nat$ and 
a sequence of actions $\alpha_1,..,\alpha_k\in\cA$ such that $P(s_{k+1,\phi}=u_2| s_{1,\phi}=u_1, a_1=\alpha_1...a_k=\alpha_k)>0$. Having that in mind, we introduce the following definition.
\begin{definition}\label{def:envir}
We say that an environment  $e$ with a state representation function $\phi$ is Markov, or, for short, that $\phi$ is 
a Markov model (of $e$), if the process $(s_{t,\phi},a_t,r_t), t\in\Nat$ is a (weakly communicating)  Markov decision process.
\end{definition}

For example, consider a state-representation function $\phi$ that depends only on the last observation, and that partitions 
the observation space into finitely many cells. Then an environment is Markov with this representation function if 
the probability distribution on the next cells only depends on the last observed cell and action.
Note that there may be many state-representation functions with which an environment $e$ is Markov.



\vspace{-2mm}
\section{Main results}\label{sec:main}

\vspace{-2mm}
Given a set $\Phi = \{ \phi_j;\, j\leq J\}$ of $J$  state-representation functions (models), 
one of which being a  Markov model of the unknown environment $e^\star$, we want to construct a strategy that performs nearly as well as the best 
algorithm that  knows which $\phi_j$ is Markov, and knows all the probabilistic characteristics (transition probabilities and rewards)
of the MDP corresponding to this model. For that purpose we define the regret of any strategy at time $T$, like in \cite{UCRL2,Bartlett2009}, as 

\vspace{-5mm}
\[
 \Delta(T) \eqdef T\rho^\star - \sum_{t=1}^Tr_t\,,
\]

\vspace{-3mm}
where $r_t$ are the rewards received when following the proposed strategy and
$\rho^\star$ is the average optimal value in the best Markov model, i.e.,
$\rho^\star = \lim_{T} \frac{1}{T}\Esp(\sum_{t=1}^Tr_t(\pi^\star))$ where $r_t(\pi^\star)$ are the rewards received when following the optimal policy for the best Markov model.
Note that this definition makes sense since when the MDP is  weakly communicating, 
the average optimal value of reward does not depend on the initial state. 
Also, one could replace $T\rho^*$ with the expected sum of rewards obtained in $T$ steps (following the optimal policy) at the price of an additional $O(\sqrt{T})$ term.

In the next subsection, we describe an algorithm that achieves a sub-linear regret of order $T^{2/3}$.

\vspace{-1mm}
\subsection{Best Lower Bound (BLB) algorithm}

\begin{figure}[hbpt]
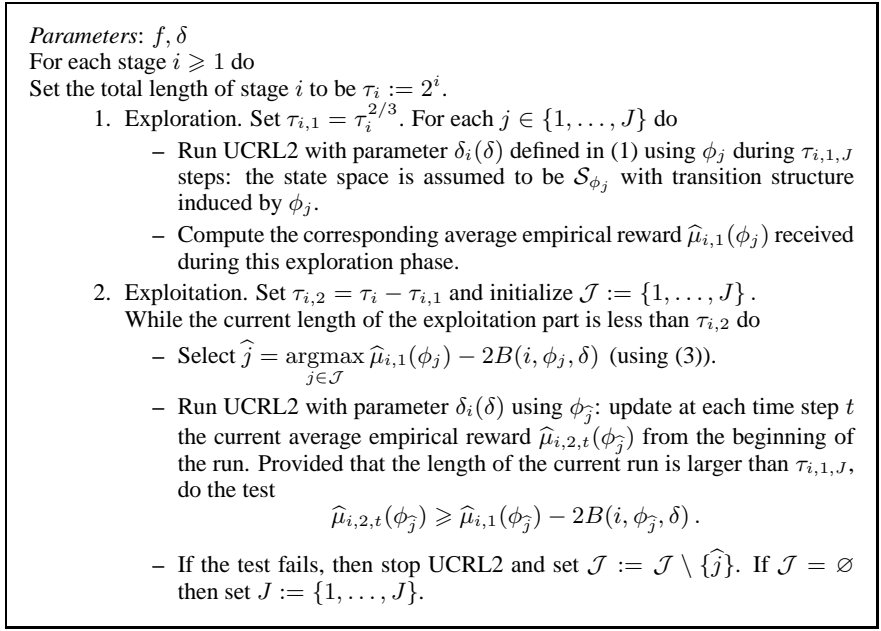

\begin{center}
\bookbox{\small
\emph{Parameters}: $f,\delta$\\ 
For each stage $i\geq 1$ do\\
Set the total length of stage $i$ to be $\tau_i :=2^i$.

\vspace{-2mm}
\begin{itemize}
\item[1.] Exploration. Set $\tau_{i,1}= \tau_i^{2/3}$. For each $j\in\{1,\dots,J\}$ do
\begin{itemize}
\item Run UCRL2 with parameter $\delta_i(\delta)$ defined in~(\ref{eq:delta}) using $\phi_j$ during $\tau_{i,1,J}$ steps:
the state space is assumed to be $\cS_{\phi_j}$ with transition structure induced by $\phi_j$.
\item Compute the corresponding average empirical reward $\hat \mu_{i,1}(\phi_j)$ received during this exploration phase.
\end{itemize}

\item[2.] Exploitation. Set $\tau_{i,2}=\tau_i- \tau_{i,1}$ and initialize $\cJ := \{1,\dots,J\}\,.$\\
While the current length of the exploitation part is less than $\tau_{i,2}$  do
\begin{itemize}
\item Select $\displaystyle{\hat j = \argmax_{j\in\cJ} \hat \mu_{i,1}(\phi_j) -  2B(i,\phi_j,\delta) \,}$ (using~(\ref{eq:confidence})).
\item Run UCRL2 with parameter $\delta_i(\delta)$ using $\phi_{\hat j}$:
update at each time step $t$ the current average empirical reward $\hat \mu_{i,2,t}(\phi_{\hat j})$ from the beginning of the run.
Provided that the length of the current run is larger than $\tau_{i,1,J}$, do the test
\[
\hat \mu_{i,2,t}(\phi_{\hat j})\geq \hat \mu_{i,1}(\phi_{\hat j}) - 2 B(i,\phi_{\hat j},\delta) \,.
\]
\item If the test fails, then stop UCRL2 and set $\cJ := \cJ \setminus \{\hat j\}$.
If $\cJ=\varnothing$ then set $J:= \{1,\dots,J\}$.
\end{itemize}

\end{itemize}
}

\vspace{-2mm}
\caption{\label{fig:Algorithm} The Best-Lower-Bound selection strategy.}
\end{center}

\vspace{-4mm}
\end{figure}

\vspace{-1mm}
In this section, we introduce the Best-Lower-Bound (BLB) algorithm, described in  Figure~\ref{fig:Algorithm}.

The algorithm works in stages of doubling length. Each stage consists in 2 phases: an exploration and an exploitation phase. In the exploration phase, BLB plays the UCRL2 algorithm on each model $(\phi_j)_{1\leq j\leq J}$ successively, as if each model $\phi_j$ was a Markov model, for a fixed number $\tau_{i,1,J}$ of rounds. The exploitation part consists in selecting first the model with highest lower bound, according to the empirical rewards obtained in the previous exploration phase. This model is initially selected for the same time as in the exploration phase, and then a test decides to either continue playing this model (if its performance during exploitation is still above the corresponding lower bound, i.e.~if the rewards obtained are still at least as good as if it was playing the best model). If it does not pass the test, then another model (with second best lower-bound) is select and played, and so on. Until the exploitation phase (of fixed length $\tau_{i,2}$) finishes and the next stage starts.

\vspace{-6mm}
The length of stage $i$ is 
fixed and defined to be $\tau_i\eqdef2^{i}$. Thus for a total time horizon $T$, 
the number of stages $I(T)$ before time $T$ is $I(T)\eqdef\llcorner\log_2(T+1)\lrcorner$. Each stage $i$ (of length $\tau_{i}$) is further decomposed into an exploration (length $\tau_{i,1}$) and an exploitation (length $\tau_{i,2}$) phases.

{\bf Exploration phase.} All the models $\{\phi_j\}_{j\leq J}$ are played one after another for the same amount of time $\tau_{i,1,J} \eqdef \frac{\tau_{i,1}}{J}$.
Each episode $1\leq j\leq J$ consists in running the UCRL2 algorithm using the model of states and transitions induced by the state-representation function $\phi_j$. Note that UCRL2 does not require the horizon $T$ in advance, but requires a parameter $p$ in order to ensure 
a near optimal regret bound with probability higher than $1-p$. We define this parameter $p$ to be $\delta_i(\delta)$ in stage $i$, where

\vspace{-7mm}
\beqa\label{eq:delta}
\delta_i(\delta) \eqdef (2^i - (J^{-1}+1)2^{2i/3}+4)^{-1}2^{-i+1}\delta\,.
\eeqa

\vspace{-3mm}
The average empirical reward received during each episode is written $\hat \mu_{i,1}(\phi_j)$.

{\bf Exploitation phase.} We use the empirical rewards $\hat \mu_{i,1}(\phi_j)$ received in the previous exploration part of stage $i$ together with
a confidence bound in order to select the model to play. Moreover, a model $\phi$ is no longer run for a fixed period of time (as in the exploration part of stage $i$), 
but for a period $\tau_{i,2}(\phi)$ that depends on some test; we first initialize $\cJ:=\{1,\dots,J\}$  and then choose
\beqa\label{eq:select}
\hat j \eqdef \argmax_{j\in\cJ} \hat \mu_{i,1}(\phi_j) - 2B(i,\phi_j,\delta)\,,
\eeqa

\vspace{-5mm}
where we define

\vspace{-8mm}
\beqa\label{eq:confidence}
 B(i,\phi,\delta) \eqdef 34 f(\tau_i-1+\tau_{i,1})|\cS_\phi| \sqrt{\frac{A\log(\frac{\tau_{i,1,J}}{\delta_i(\delta)})}{\tau_{i,1,J}}}\,,
\eeqa

\vspace{-3mm}
where $\delta$ and the function $f$ are parameters of the BLB algorithm.
Then UCRL2 is played using the selected model $\phi_{\hat j}$ for the parameter $\delta_i(\delta)$.
In parallel we test whether the average empirical reward we receive
during this exploitation phase is high enough; at time $t$, if the length of the current episode is larger than $\tau_{1,i,J}$, we test if
\beqa\label{eq:test}
 \hat \mu_{i,2,t}(\phi_{\hat j})\geq \hat \mu_{i,1}(\phi_{\hat j}) - 2B(i,\phi_{\hat j},\delta).
\eeqa
If the test is positive, we keep playing UCRL2 using the same model. Now, if the test fails, then the model $\hat j$ is discarded (until the end of stage $i$) i.e.~we update $\cJ:= \cJ \setminus \{\hat j\}$ and we select a new one according to \eqref{eq:select}.
We repeat those steps until the total time $\tau_{i,2}$ of the exploitation phase of stage $i$ is over.

{\bf Remark}
 Note that the model selected for exploitation in~\eqref{eq:select} is the one that has the best lower bound. This is a pessimistic (or robust) selection strategy. We know that if the right model is selected, then with high probability, this model will be kept during the whole exploitation phase. If this is not the right model, then either the policy provides good rewards and we should keep playing it, or it does not, in which case it will not pass the test~\eqref{eq:test} and will be removed from the set of models that will be exploited in this phase.

\vspace{-2mm}
\subsection{Regret analysis}\label{sec:regret}

\vspace{-1mm}
\begin{theorem}[Main result]\label{thm:main}
Assume that a finite set of $J$ state-representation functions $\Phi$ is given, and there exists at least one 
function $\phi^\star\in\Phi$ such that with $\phi^\star$ as a state-representation function  the environment is a Markov decision process.
If there are several such models, let $\phi^\star$ be the one with the highest average reward of the optimal policy of the corresponding MDP.
Then the regret (with respect to the optimal policy corresponding to $\phi^*$) of the BLB algorithm 
run with parameter $\delta$, for any horizon $T$, with probability higher than $1-\delta$ is bounded as follows
{\small
\beq\label{eq:main}
 \Delta(T) \leq cf(T)S\Big(AJ\log\big((J\delta)^{-1}\big)\log_2(T)\Big)^{1/2}T^{2/3}
+c'DS\Big(A\log(\delta^{-1})\log_2(T)T\Big)^{1/2} + c(f,D),
\eeq}
for some numerical constants $c,c'$ and $c(f,D)$.
The parameter $f(t)$ can be chosen to be any increasing function, for instance the choice $f(t):=\log_2 t +1$, gives $c(f,D) \leq 2^{D}$.
\end{theorem}

\vspace{-3mm}
The proof of this result is reported in Section~\ref{sec:proof}.

{\bf Remark.}
Importantly, the algorithm considered here \textit{does not} know in advance the diameter $D$ of the true model, nor the time horizon $T$.
Due to this lack of knowledge, it uses a guess $f(t)$ (e.g. $\log(t)$) on this diameter, which result in 
the additional regret term $c(f,D)$ and the additional factor $f(T)$; knowing $D$ would enable to remove both of them, but this 
is a strong assumption. 
Choosing $f(t):=\log_2 t +1$ gives a bound which is of order $T^{2/3}$ in $T$ but is exponential in $D$; taking 
$f(t):=t^\epsilon$ we get a bound of order $T^{2/3+\epsilon}$ in $T$ but of polynomial order $1/\epsilon$ in $D$. 

\vspace{-2mm}
\section{Discussion and outlook}\label{sec:disc}

\vspace{-2mm}
{\bf Intuition.}
The main idea why this algorithm works is as follows. The ``wrong'' models are used  during 
exploitation stages only as long as they are giving rewards that are higher than the rewards that could 
be obtained in the ``true'' model.   All the models are explored sufficiently long so as to be able to estimate
the optimal reward level in the true model, and to learn its policy. 
Thus, nothing has to be known about the ``wrong'' models. 
This is in stark contrast to the usual situation in mathematical statistics, where to be able 
to test a hypothesis about a model (e.g., that the data is generated by a certain model versus 
some alternative models), one has to make assumptions about alternative models. This has to be done
in order to make sure that the Type II error is small (the power of the test is large): that this
error is small has to be proven under the alternative. Here, although we are solving seemingly the same problem, 
the role of the Type II error is played 
by the rewards. As long as the rewards are high we do not care where the model we are using is correct or not.
We only have to ensure that the true model passes the test.

{\bf Assumptions.} A crucial assumption made in this work is that the ``true'' model $\phi^*$ belongs to a known finite set.
While passing from a finite to a countably infinite set appears rather straightforward, getting rid 
of the assumption that this set \textit{contains} the true model seems more difficult. What one would want to 
obtain in this setting is sub-linear  regret with respect to the performance of the optimal policy 
in the best model; this, however, seems difficult without additional assumptions on the probabilistic characteristics
of the models. Another approach not discussed here would be to try to \textit{build} a good state representation function,
as what is suggested for instance in \cite{Hutter09}.
Yet another interesting generalization in this direction would be to consider uncountable (possibly parametric but general) sets
of models. This, however, would necessarily require some heavy assumptions on the set of models. 

{\bf Regret.} The reader familiar with adversarial bandit literature will notice that our bound of order $T^{2/3}$ is worse 
than $T^{1/2}$ that usually appears in this context (see, for example \cite{exp3}). The reason is that our notion of regret is 
different: in adversarial bandit literature, the regret is measured with respect to the best choice of the arm  {\em for the given 
fixed history}. In contrast, we measure the regret with respect to the best policy (for knows the correct model and its parameters) 
that, in general, would obtain completely different (from what our algorithm would get) rewards and observations right from the beginning.
 
{\bf Estimating the diameter?} As previously mentioned, a possibly large additive constant $c(f,D)$ appears in the regret since we do not known a bound on the diameter of the MDP in the ``true'' model, and use $\log T$ instead.
Finding a way to properly address this problem by estimating online the diameter of the MDP is an interesting open question.
Let us provide two intuitions concerning this problem.
First, we notice that, as reported in \cite{UCRL2}, when we compute an optimistic model based on the empirical rewards and transitions of the true model,
the span of the corresponding optimistic value function $sp(\hat V^+)$ is always smaller than the diameter $D$. This span increases as we get more rewards and transitions samples, which gives a natural empirical lower bound on $D$.
However, it seems quite difficult to compute a tight empirical upper bound on $D$ (or $sp(\hat V^+)$).
In \cite{Bartlett2009}, the authors derive a regret bound that scales with the span of the true value function $sp(V^\star)$, which is also less than $D$, and can be significantly smaller in some cases.
However, since we do not have the property that $sp(\hat V^+)\leq sp(V^\star)$, we need to introduce an explicit penalization in order to control the span of the computed optimistic models, and this requires assuming we know an upper bound $B$ on $sp(V^\star)$ in order to guarantee a final regret bound scaling with $B$. Unfortunately this does not solve the estimation problem of $D$, which remains an open question.


\vspace{-2mm}
\section{Proof of Theorem~\ref{thm:main}}\label{sec:proof}

\vspace{-2mm}
In this section, we now detail the proof of Theorem~\ref{thm:main}.
The proof is stated in several parts.
First we remind a general confidence bound for the UCRL2 algorithm in the true model.
Then we decompose the regret into the sum of the regret in each stage $i$.
After analyzing the contribution to the regret in stage $i$, we then
gather all stages and tune the length of each stage and episode in order to get the final regret bound.

\vspace{-1mm}
\subsection{Upper and Lower confidence bounds}

\vspace{-3mm}
From the analysis of UCRL2 in \cite{UCRL2}, we have the property that
with probability higher than $1-\delta'$, the regret of UCRL2 when run for $\tau$ consecutive many steps from time $t_1$ in the true model $\phi^\star$ is upper bounded by

\vspace{-9mm}
\beqa
\rho^\star - \frac{1}{\tau}\sum_{t=t_1}^{t_1+\tau-1} r_t \leq 34 D|\cS_{\phi^\star}| \sqrt{\frac{A\log(\frac{\tau}{\delta'})}{\tau}}\,,
\eeqa

\vspace{-3mm}
where $D$ is the diameter of the MDP. 
What is interesting is that this diameter does not need to be known by the algorithm.
Also by carefully looking at the proof of UCRL, it can be shown that the following bound is also valid with probability higher than $1-\delta'$:

\vspace{-4mm}
\[ 
\frac{1}{\tau}\sum_{t=t_1}^{t_1+\tau-1} r_t -\rho^\star \leq 34 D|\cS_{\phi^\star}| \sqrt{\frac{A\log(\frac{\tau}{\delta'})}{\tau}}\,.
\]

\vspace{-2mm}
We now define the following quantity, for every model $\phi$, episode length $\tau$ and $\delta'\in(0,1)$

\vspace{-6mm}
\beqa
 B_{D}(\tau,\phi,\delta') \eqdef 34 D|\cS_\phi| \sqrt{\frac{A\log(\frac{\tau}{\delta'})}{\tau}}\,.
\eeqa

\vspace{-1mm}
\subsection{Regret of stage $i$}

\vspace{-3mm}
 In this section we analyze the regret of the stage $i$, which we denote $\Delta_i$.
Note that since each stage $i\leq I$ is of length $\tau_i = 2^i$ except the last one $I$ that may stop before, 
we have 

\vspace{-7mm}
\beqa
 \Delta(T) = \sum_{i=1}^{I(T)}\Delta_i\,,
\eeqa

\vspace{-3mm}
where $I(T) = \llcorner\log_2(T+1)\lrcorner$. 
We further decompose $\Delta_i=\Delta_{1,i} + \Delta_{i,2}$ into the regret corresponding to the exploration stage $\Delta_{1,i}$
and the regret corresponding to the exploitation stage $\Delta_{i,2}$.

Recall that $\tau_{i,1}$ is the total length of the exploration stage $i$
and $\tau_{i,2}$ is the total length of the exploitation stage $i$.
Then for each model $\phi$, we write 
$\tau_{i,1,J}\eqdef\frac{\tau_{i,1}}{J}$ the number of consecutive steps during which the UCRL2 algorithm is run with model $\phi$
in the exploration stage $i$, and $\tau_{i,2}(\phi)$ the number of consecutive steps during which the UCRL2 algorithm is run with model $\phi$
in the exploitation stage~$i$.

{\bf Good and Bad models.}
Let us now introduce the two following sets of models, 
defined after the end of the exploration stage, i.e. at time $t_i$.
\beqan
 \cG_i &\eqdef& \{ \phi \in \Phi\,\,;\,\, \hat \mu_{i,1}(\phi) - 2 B(i,\phi,\delta) \ge  \hat \mu_{i,1}(\phi^\star) - 2 B(i,\phi^\star,\delta) \}\backslash\{\phi^*\}\,,\\
 \cB_i &\eqdef& \{ \phi \in \Phi\,\,;\,\, \hat \mu_{i,1}(\phi) - 2 B(i,\phi,\delta) <  \hat \mu_{i,1}(\phi^\star) - 2 B(i,\phi^\star,\delta) \}\,.
\eeqan

\vspace{-3mm}
With this definition, we have the decomposition $\Phi= \cG_i \cup \{ \phi^\star \} \cup\cB_i$.

\vspace{-1mm}
\subsubsection{Regret in the exploration phase}
%
%

\vspace{-2mm}
Since in the exploration stage $i$ each model $\phi$ is run for $\tau_{i,1,J}$ many steps,
the regret for each model $\phi\neq\phi^\star$ is bounded by $\tau_{i,1,J}\rho^\star$.
Now the regret for the true model is $\tau_{i,1,J}(\rho^\star - \hat \mu_1(\phi^\star))$, 
thus the total contribution to the regret in the exploration stage $i$ is upper-bounded by
\beqa
\Delta_{i,1} \leq \tau_{i,1,J}(\rho^\star - \hat \mu_1(\phi^\star)) + (J-1)\tau_{i,1,J}\rho^\star\,.
\eeqa

\subsubsection{Regret in the exploitation phase}

\vspace{-2mm}
By definition, all models in $\cG_i\cup\{\phi^\star\}$ are selected 
before any model in $\cB_i$ is selected.

{\bf The good models.}
Let us consider some $\phi\in\cG_i$ and an event $\Omega_i$ under which the exploitation phase does not reset.
The test (equation \eqref{eq:test}) starts after $\tau_{i,1,J}$, thus, since there is not reset, either $\tau_{i,2}(\phi)=\tau_{i,1,J}$ 
in which case the contribution to the regret is  bounded by $\tau_{i,1,J}\rho^\star\,,$
or $\tau_{i,2}(\phi)>\tau_{i,1,J}$, in which case the regret during the $(\tau_{i,2}(\phi)-1)$ steps (where the test was successful) is bounded by
\beqan
(\tau_{i,2}(\phi)-1)(\rho^\star - \hat \mu_{i,2,\tau_{i,2}(\phi)-1}(\phi))&\leq& 
(\tau_{i,2}(\phi)-1)(\rho^\star - \hat \mu_{i,1}(\phi)+2B(i,\phi,\delta))\\
&\leq&(\tau_{i,2}(\phi)-1)(\rho^\star - \hat \mu_{i,1}(\phi^\star) + 2B(i,\phi^\star,\delta))\,,
\eeqan
 and now since in the last step $\phi$ fails to pass the test, 
this adds a  contribution to the regret at most $\rho^\star$.

We deduce that the total contribution to the regret of all the models $\phi\in\cG_i$ in the exploitation stages on the event $\Omega_i$ is bounded by
\beqa
 \Delta_{i,2}(\cG_i)\leq\sum_{\phi\in\cG} \max\{\tau_{i,1,J}\rho^\star,(\tau_{i,2}(\phi)-1)(\rho^\star - \hat \mu_{i,1}(\phi^\star) + 2B(i,\phi^\star,\delta)) + \rho^\star  \}\,.
\eeqa

\vspace{-3mm}
{\bf The true model.}
First, let us note that since the total regret of the true model during the exploitation step $i$ is given by

\vspace{-6mm}
\[
 \tau_{i,2}(\phi^\star)(\rho^\star - \hat \mu_{i,2,t}(\phi^\star))\,,
\]
then the total regret of the exploration and exploitation stages in episode $i$ on $\Omega_i$ is bounded by
{\small
\beqan
\Delta_i &\leq& \tau_{i,1,J}(\rho^\star - \hat \mu_1(\phi^\star)) + \tau_{i,1,J}(J-1)\rho^\star+ 
\tau_{i,2}(\phi^\star)(\rho^\star - \hat \mu_{i,2,t_i+\tau_{i,2}}(\phi^\star))+\\
&&\sum_{\phi\in\cG_i} \max\{\tau_{i,1,J}\rho^\star,(\tau_{i,2}(\phi)-1)(\rho^\star - \hat \mu_{i,1}(\phi^\star) + 2B(i,\phi^\star,\delta)) + \rho^\star  \} 
+ \sum_{\phi\in\cB_i} \tau_{i,2}(\phi)\rho^\star\,.
\eeqan}

\vspace{-3mm}
Now from the analysis provided in \cite{UCRL2} we know that when we run the UCRL2 with the true model $\phi^\star$ with parameter $\delta_i(\delta)$, 
then there exists an event $\Omega_{1,i}$ of probability at least $1-\delta_i(\delta)$ such that on this event

\vspace{-7mm}
\[\rho^\star - \hat \mu_{i,1}(\phi^\star) \leq B_{D}(\tau_{i,1,J},\phi^\star,\delta_i(\delta))\,,\]
and similarly there exists an event $\Omega_{2,i}$ of probability  at least $1-\delta_i(\delta)$, such that on this event

\vspace{-5mm}
\[\rho^\star - \hat \mu_{i,2,t}(\phi^\star) \leq B_{D}(\tau_{i,2}(\phi^\star),\phi^\star,\delta_1(\delta))\,.\]
Now we show that, with high probability, 
the true model $\phi^\star$ passes all the tests (equation \eqref{eq:test}) until the end of the episode $i$,
and thus equivalently, with high probability no model $\phi\in\cB_i$ is selected,  so that
$\displaystyle{\sum_{\phi\in\cB_i}\tau_{i,2}(\phi) = 0}$.

\vspace{-2mm}
For the true model, after $\tau(\phi^\star,t)\geq\tau_{i,1,J}$,
there remains at most $(\tau_{i,2} - \tau_{i,1,J}+1)$ possible timesteps where we do the test for the true model $\phi^\star$.
For each test we need to control $\mu_{i,2,t}(\phi^\star)$, and the event corresponding to $\hat \mu_{i,1}(\phi^\star)$ is shared by all the tests.
Thus we deduce that with probability higher than $1-(\tau_{i,2} - \tau_{i,1,J}+2)\delta_i(\delta)$ we have simultaneously 
on all time step until the end of exploitation phase of stage $i$,
\beqan
 \hat \mu_{i,2,t}(\phi^\star) - \hat \mu_{i,1}(\phi^\star) &=&  \hat \mu_{i,2,t}(\phi^\star) - \rho^\star + \rho^\star - \hat \mu_{i,1}(\phi^\star)\\
&\geq& - B_{D}(\tau(\phi^\star,t),\phi^\star,\delta_i(\delta))  - B_{D}(\tau_{i,1,J},\phi^\star,\delta_i(\delta)) \\
&\geq& - 2B_{D}(\tau_{i,1,J},\phi^\star,\delta_i(\delta))\,.
\eeqan

Now provided that $f(t_i)\geq D$, then $\displaystyle{B_{D}(\tau_{i,1,J},\phi^\star,\delta_i(\delta))\leq B(i,\phi^\star,\delta)\,,}$
thus the true model passes all tests until the end of the exploitation part 
of stage $i$ on an event $\Omega_{3,i}$ of probability higher than $1-(\tau_{i,2} - \tau_{i,1,J}+2)\delta_i(\delta)$.
Since there is no reset, we can choose $\Omega_i\eqdef\Omega_{3,i}$. Note that on this event, we thus have $\displaystyle{\sum_{\phi\in\cB_i}\tau_{i,2}(\phi) = 0}$.

By using a union bound over the events $\Omega_{1,i},\Omega_{2,i}$ and $\Omega_{3,i}$, 
then we deduce that with probability higher than $1-(\tau_{i,2} - \tau_{i,1,J}+4)\delta_i(\delta)$, 
\beqan
\Delta_i &\leq& \tau_{i,1,J}B_{D}(\tau_{i,1,J},\phi^\star,\delta_i(\delta))) + 
[\tau_{i,1,J}(J-1) + |\cG_i|]\rho^\star+  \tau_{i,2}(\phi^\star)B_{D}(\tau_{i,2}(\phi^\star),\phi^\star,\delta_i(\delta))\\
&& +\sum_{\phi\in\cG_i} \max\{(\tau_{i,1,J}-1)\rho^\star,(\tau_{i,2}(\phi)-1)(B_D(\tau_{i,1,J},\phi^\star,\delta_i(\delta))+2B(i,\phi^\star,\delta)\}\,.
\eeqan

Now using again the fact that $f(t_i) \geq D$, and after some simplifications, we deduce that
\beqan
\Delta_i&\leq&\tau_{i,1,J}B_{D}(\tau_{i,1,J},\phi^\star,\delta_i(\delta)) +\tau_{i,2}(\phi^\star)B_{D}(\tau_{i,2}(\phi^\star),\phi^\star,\delta_i(\delta))\\ 
&&+ \sum_{\phi\in\cG_i}(\tau_{i,2}(\phi)-1) 3B(i,\phi^\star,\delta)+\tau_{i,1,J}(J+|\cG_i|-1)\rho^\star\,.
\eeqan

Finally, we use the fact that $\tau B_D(\tau,\phi^\star,\delta_i(\delta))$ is increasing with $\tau$
to deduce the following rough bound that holds with probability higher than $1-(\tau_{i,2} - \tau_{i,1,J}+4)\delta_i(\delta)$ 
\beqan
\Delta_i&\leq&\tau_{i,2}B(i,\phi^\star,\delta) + \tau_{i,2}B_{D}(\tau_{i,2},\phi^\star,\delta_i(\delta))
+ 2J\tau_{i,1,J}\rho^\star\,,
\eeqan
where we used the fact that $\displaystyle{\tau_{i,2} = \tau_{i,2}(\phi^\star) + \sum_{\phi\in\cG}\tau_{i,2}(\phi)}\,.$

\vspace{-2mm}
\subsection{Tuning the parameters of each stage.}

\vspace{-2mm}
We now conclude by tuning the parameters of each stage, i.e. the probabilities $\delta_i(\delta)$ and the 
length $\tau_i$, $\tau_{i,1}$ and $\tau_{i,2}$.
The total length of stage $i$ is by definition
\[\tau_i=\tau_{i,1}+\tau_{i,2}=\tau_{i,1,J}J+\tau_{i,2}\,,\] 
where $\tau_i=2^{i}\,.$
So we set $\tau_{i,1} \eqdef \tau_{i}^{2/3}$ and then we have $\tau_{i,2} \eqdef \tau_{i} - \tau_{i}^{2/3}$ and 
$\tau_{i,1,J} = \frac{\tau_{i}^{2/3}}{J}$.
Now using these values and the definition of the bound $B(i,\phi^\star,\delta)$, and $B_{D}(\tau_{i,2},\phi^\star,\delta_i(\delta))$,
we deduce with probability higher than $1-(\tau_{i,2} - \tau_{i,1,J}+4)\delta_i(\delta)$  the following upper bound
\beqan
\Delta_i&\leq &34f(t_i)S\sqrt{AJ\log\Big(\frac{\tau_i^{2/3}}{J\delta_i(\delta)}\Big)}\tau_i^{2/3} + 
34DS\sqrt{A\log\Big(\frac{\tau_i}{\delta_i(\delta)}\Big)\tau_i}
+2\tau_{i}^{2/3}\rho^\star\,,
\eeqan

\vspace{-4mm}
with $t_i = 2^i-1 + 2^{2i/3}$ and where we used the fact that $\Big(\frac{J}{\tau_i^{2/3}}\Big)^{1/2}\tau_{i,2} \leq \sqrt{J}\tau_i^{2/3}$.

We now define $\delta_i(\delta)$ such that $\displaystyle{\delta_i(\delta) \eqdef (2^i - (J^{-1}+1)2^{2i/3}+4)^{-1}2^{-i+1}\delta\,.}$

\vspace{-2mm}
Since for the stages $i\in \cI_0\eqdef \{ i\geq 1 ; f(t_i)<D\}$, the regret is bounded by $\Delta_i\leq\tau_i\rho^\star$, 
then the total cumulative regret of the algorithm is bounded with probability higher than $1-\delta$ (using the defition of the $\delta_i(\delta)$) by

\vspace{-6mm}
\[
 \Delta(T) \leq 
\sum_{i\notin\cI_0}[34f(t_{i})S\sqrt{JA\log\Big(\frac{2^{8i/3}}{J\delta}\Big)}+2]2^{2i/3} + 
34DS\sqrt{A\log\Big(\frac{2^{3i}}{\delta}\Big)2^{i}}+ \sum_{i\in\cI_0} 2^i\rho^\star \,.
\]
where $t_i = 2^i-1 + 2^{2i/3}\leq T$.

We conclude by using the fact that since $I(T) \leq \log_2(T+1)$, then
with probability higher than $1-\delta$, the following bound on the regret holds 
\[
 \Delta(T) \leq cf(T)S\Big(AJ\log(J\delta)^{-1}\log_2(T)\Big)^{1/2}T^{2/3}
+c'DS\Big(A\log(\delta^{-1})\log_2(T)T\Big)^{1/2} + c(f,D)\,.
\]
for some constant $c,c'$, and where $c(f,D) = \sum_{i\in\cI_0} 2^i\rho^\star$.
Now for the special choice when $f(T) \eqdef \log_2(T+1)$, then $i\in\cI_0$ means $2^i+ 2^{2i/3}<2^D+2$, thus we must have $i<D$, 
and thus $c(f,d) \leq 2^{D}$.

\newpage
\subsection*{Acknowledgements}
This research
was partially supported by the French Ministry of Higher Education and Research, Nord-
Pas-de-Calais Regional Council and FEDER through CPER 2007-2013, ANR projects EXPLO-RA
(ANR-08-COSI-004) and Lampada (ANR-09-EMER-007), by the European Community’s Seventh
Framework Programme (FP7/2007-2013) under grant agreement 231495 (project CompLACS), and
by Pascal-2.
{
\bibliography{bib_MsRl}
\bibliographystyle{plain}
}

\end{document}